\documentclass[10pt,twocolumn,letterpaper]{article}

\usepackage[pagenumbers]{cvpr} % To force page numbers, e.g. for an arXiv version

\usepackage{algorithm}
\usepackage{algorithmic}
\usepackage{booktabs}
\usepackage{multirow}
\usepackage{amssymb}
\usepackage{amsmath,amsfonts,bm}

\def\eqref#1{equation~\ref{#1}}
\def\1{\bm{1}}

\def\eps{{\epsilon}}

\def\vc{{\bm{c}}}
\def\vd{{\bm{d}}}
\def\ve{{\bm{e}}}

\def\vg{{\bm{g}}}
\def\vh{{\bm{h}}}

\def\vm{{\bm{m}}}
\def\vn{{\bm{n}}}

\def\vp{{\bm{p}}}

\def\mC{{\bm{C}}}

\def\mE{{\bm{E}}}

\def\mI{{\bm{I}}}

\def\mK{{\bm{K}}}

\def\mN{{\bm{N}}}
\def\mO{{\bm{O}}}

\def\mQ{{\bm{Q}}}

\def\mS{{\bm{S}}}

\def\mV{{\bm{V}}}
\def\mW{{\bm{W}}}
\def\mX{{\bm{X}}}

\DeclareMathAlphabet{\mathsfit}{\encodingdefault}{\sfdefault}{m}{sl}
\SetMathAlphabet{\mathsfit}{bold}{\encodingdefault}{\sfdefault}{bx}{n}

\def\gG{{\mathcal{G}}}

\def\gL{{\mathcal{L}}}

\def\gN{{\mathcal{N}}}

\newcommand{\R}{\mathbb{R}}

\DeclareMathOperator*{\Ours}{\textbf{DMRA}}
\DeclareMathOperator*{\encodernode}{Encoder_{Node}}
\DeclareMathOperator*{\encoderedge}{Encoder_{Edge}}
\DeclareMathOperator*{\Linear}{Linear}
\usepackage{xcolor}
\usepackage{colortbl}
\definecolor{cvprblue}{rgb}{0.21,0.49,0.74}
\usepackage[pagebackref,breaklinks,colorlinks,allcolors=cvprblue]{hyperref}

\def\paperID{} % *** Enter the Paper ID here
\def\confName{CVPR}
\def\confYear{2025}

\title{Diffusion Model with Representation Alignment for Protein Inverse Folding}

\def\spaces{~~~~~}
\author{Chenglin Wang\textsuperscript{1}\footnotemark[1]\spaces{}
Yucheng Zhou\textsuperscript{2}\footnotemark[1]\spaces{}
Zijie Zhai\textsuperscript{1}\spaces{}
Jianbing Shen\textsuperscript{2}\spaces{}
Kai Zhang\textsuperscript{1}\footnotemark[2]\spaces{}\\\\
\textsuperscript{1}East China Normal University \spaces
\textsuperscript{2}SKL-IOTSC, CIS, University of Macau \\
{\tt\small 52275901013@stu.ecnu.edu.cn \spaces yucheng.zhou@connect.um.edu.mo}
}

\begin{document}
\maketitle
\renewcommand{\thefootnote}{\fnsymbol{footnote}}
\footnotetext[1]{Equal Contribution.} \footnotetext[2]{Corresponding Author.}

\begin{abstract}
Protein inverse folding is a fundamental problem in bioinformatics, aiming to recover the amino acid sequences from a given protein backbone structure. 
Despite the success of existing methods, they struggle to fully capture the intricate inter-residue relationships critical for accurate sequence prediction.
We propose a novel method that leverages diffusion models with representation alignment ($\Ours$), which enhances diffusion-based inverse folding by 
 (1) proposing a shared center that aggregates contextual information from the entire protein structure and selectively distributes it to each residue; and 
(2) aligning noisy hidden representations with clean semantic representations during the denoising process. This is achieved by predefined semantic representations for amino acid types and a representation alignment method that utilizes type embeddings as semantic feedback to normalize each residue.
In experiments, we conduct extensive evaluations on the CATH4.2 dataset to demonstrate that $\Ours$ outperforms leading methods, achieving state-of-the-art performance and exhibiting strong generalization capabilities on the TS50 and TS500 datasets.  
\end{abstract}

\section{Introduction}
Protein inverse folding, a crucial task in bioinformatics and computational biology, aims to reversely explore possible amino acid (AA) sequences from a given protein 3D structure~\cite{o2018spin2,ingraham2019generative,yue1992inverse}. 
These predicted sequences can autonomously fold into functional proteins, enabling the design of novel proteins with desired structural and functional properties. 
Moreover, some of these designed proteins, which may not occur naturally, have significant applications in biological research, including drug design and antibody engineering~\cite{khoury2014protein,dauparas2022robust,watson2023novo}.

Recent studies have revealed the effective application of neural networks in analyzing protein 3D structures~\cite{sverrisson2021fast, quan2024clustering, chen2023fff}. 
Predicting AA sequences based on protein backbone structures is a 3D structure-to-sequence mapping problem.
Numerous studies have utilized GNNs \cite{EGNN,PaiNN} to extract protein structural features (i.e., residue features and their connections)~\cite{ingraham2019generative,dauparas2022robust,GVP21JingESTD}, followed by the Transformer architecture to generate protein sequences in an autoregressive manner ~\cite{vaswani2017attention,devlin2018bert}. 
\begin{figure}[!t]
    \centering
    \includegraphics[width=\linewidth]{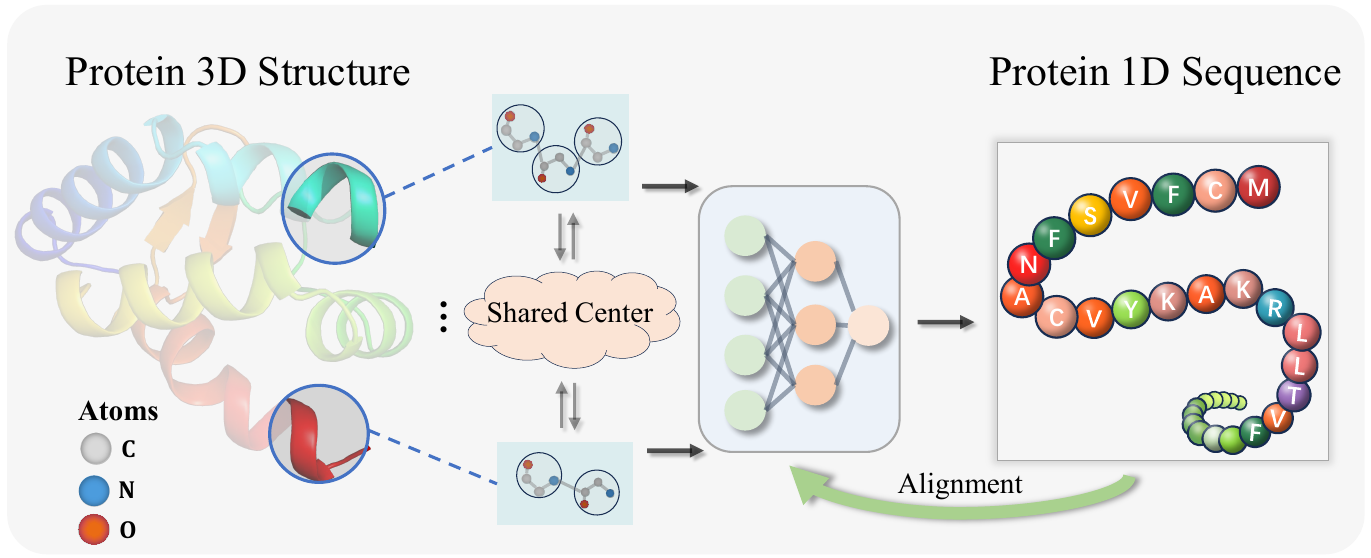}
    \caption{The protein inverse folding problem defines the mapping from a 3D structure to an amino acid sequence. “Shared Center” refers to the shared information among all residues in the protein chain, which is used to maintain the communication of residues. Each residue on the chain consists of three atoms: $\mC$, $\mN$, and $\mO$, with the central carbon atom known as $\mC_\alpha$. The characters on the right represent different types of amino acids.}
    \label{fig:intro}
    \vspace{-0.4cm}
\end{figure}

Recently,  diffusion models have been extensively applied for generating meaningful contents in both vision and language~\cite{ho2020denoising,peebles2023scalable,rombach2022high,zhang2023adding,austin2021structured,ardiff,diffuseq}, due to their ability to produce highly diverse yet faithful data from the desired distribution. 
Moreover, diffusion models have shown promise in analyzing and interpreting protein structures. 
For instance, DPLM~\cite{wang2024diffusion} adopted a discrete diffusion framework to train protein sequences, exhibiting the potential of the diffusion model for protein representation learning. 
Similarly, Grade-IF~\cite{YiZ0L023} proposed a graph diffusion model for protein inverse folding, effectively learning latent protein representations by capturing inter-residue interactions, which encapsulate various reasonable sequences for a given backbone structure.

From a biological standpoint, protein neighbor inter-residue interactions occur among spatially close amino acids and contribute to salient substructures, which is beneficial for protein representation learning~\cite{aurora1997local, toal2014local, rackovsky2010global}. 
In addition, the state of a protein chain is intrinsically linked to the collective contributions of its residues~\cite{rackovsky2010global, scheraga2016global, scheraga2014homolog}. Viewing a protein chain as a steady-state system, each residue is vital for maintaining overall stability. Therefore, effective communication among residues is essential for protein representation learning.
Recent studies improve diffusion models by aligning external representations, i.e., facilitating the alignment of noisy hidden representations with external ones during the denoising process~\cite{chen2024deconstructing, xiang2023denoising, yu2024representation}.
In AA sequences, AA types are not merely discrete tokens but also could serve as alignment guidance for residue semantics. However, existing methods mainly utilize these sequences for model supervision, neglecting their potential to provide semantic feedback for hidden representations. 

In this work, we propose a novel method that leverages diffusion models with representation alignment ($\Ours$) for protein inverse folding.
As shown in Figure~\ref{fig:intro}, to facilitate effective communication across residues on the protein chain, we propose a shared center that aggregates contextual information from the entire protein structure, distributing it selectively to each residue.
Our empirical analysis shows that accessing contextual information significantly improves the non-linear transformation capabilities of each layer, thereby improving residue representation.
Moreover, to align the noisy hidden representation with the clean semantic during the denoising process, we predefine the semantic representations for AA types and employ a representation alignment method that uses type embeddings as semantic feedback to normalize each residue. 
By integrating type-specific semantic embeddings as supplementary prior knowledge, the hidden representations of residues are better aligned with type semantics during denoising, enhancing the recovery of the AA sequence.
Finally, unlike previous methods that concatenate node and edge features for sequence prediction, we propose a cell module that more effectively integrates these features by estimating the relevance between adjacent nodes and edges.

Extensive experiments demonstrate that our method achieves state-of-the-art performance on the CATH4.2 dataset~\cite{ingraham2019generative}. To assess the generalization capacity of our approach, we also evaluate it on the TS50 and TS500 datasets~\cite{li2014ts500} and observe its significant improvements against baseline methods. Finally, we provide detailed visualization and analysis in case studies to illustrate the effectiveness of our approach.

The contributions of this work are summarized below: 
\begin{itemize} 
\item We introduce a novel diffusion model with representation alignment for protein inverse folding. 
\item We design a shared center to integrate and distribute information among residues, thereby enhancing residue representation learning. 
\item We propose a representation alignment method for AA type embeddings and noisy hidden representations during the denoising process. 
\item Our method surpasses state-of-the-art approaches on the CATH4.2 dataset and demonstrates strong generalization on the TS50 and TS500 datasets. 
\end{itemize}

\section{Related Work}
Protein inverse folding can be formulated as a structure-based conditional generation, where 3D structure can be encoded to a knn-graph. residues and their relationships can be compressed into node features and edge features.
GraphTrans~\cite{ingraham2019generative} extracted protein backbone features as structural representation, including angles and distances, etc., and then used the decoder to obtain the amino acid sequences in an autoregressive manner.
Recent works enhanced graph features to better compress structural information. GVP-GNN~\cite{GVP21JingESTD} introduced geometric vector perceptrons to replace MLP, simultaneously leveraging the geometric and relational features.
ProteinMPNN~\cite{dauparas2022robust} introduced distances between $\mN$, $\mC_\alpha$, $\mC$, $\mO$, and a virtual $\mC_\beta$ as additional input features, resulting in a significant performance.
PiFold~\cite{Gao0L23} introduced a novel residue featurizer that used virtual atoms to capture hidden information, and designed PiGNN layers to generate protein sequences.
VFN~\cite{VFN} established a set of virtual atoms for each residue and utilized a vector field operator to extract geometric features to augment the features of vector field network layers.
Besides, to fully consider sequence information in the mapping process, ESM-IF~\cite{hsu2022learning} augmented training data and used this additional data to train, resulting in significant improvements. 
LM-Design~\cite{ZhengDX0YG23} and KWDesign~\cite{gao2023KD} employed pre-trained language models to refine amino acid sequences iteratively.
These methods have achieved significant success in sequence recovery, but they may encounter challenges in encapsulating the diversity of feasible protein sequences.

In recent years, generative models have garnered significant attention~\cite{VAE, GAN, score}.DDPM~\cite{ho2020denoising} utilized the Diffusion paradigm by progressively introducing Gaussian noise into images and learning its reverse process, which has achieved remarkable success in image generation. Furthermore, Latent Diffusion~\cite{rombach2022high} and ControlNet~\cite{zhang2023adding} enhanced controllability by incorporating text as a condition for image generation. In Addition, D3PM~\cite{austin2021structured}  has extended the multinomial diffusion model by~\citet{hoogeboom2021argmax} to handle discrete data. In the realm of protein, DPLM~\cite{wang2024diffusion} employed the Diffusion framework to model protein sequences in an unconditional generation manner, leading to a better understanding of proteins. 
\citet{ZhangXLCD023} proposed the DiffPreT approach to pre-train a protein encoder by sequence-structure joint diffusion modeling and enhance DiffPreT by SiamDiff, a method to capture the correlation between different conformers of a protein.
\citet{yutong2024} proposed CPDiffusion-SS, a latent graph diffusion model that generates protein sequences based on coarse-grained secondary structural information, enhancing the reliability and diversity of the generated proteins.
GRADE-IF~\cite{YiZ0L023} proposed an innovative graph denoising diffusion model for structure-based protein sequence design, demonstrating significant potential in generating diverse protein sequences.

\begin{figure*}[!t]
    \centering
    \includegraphics[width=\linewidth]{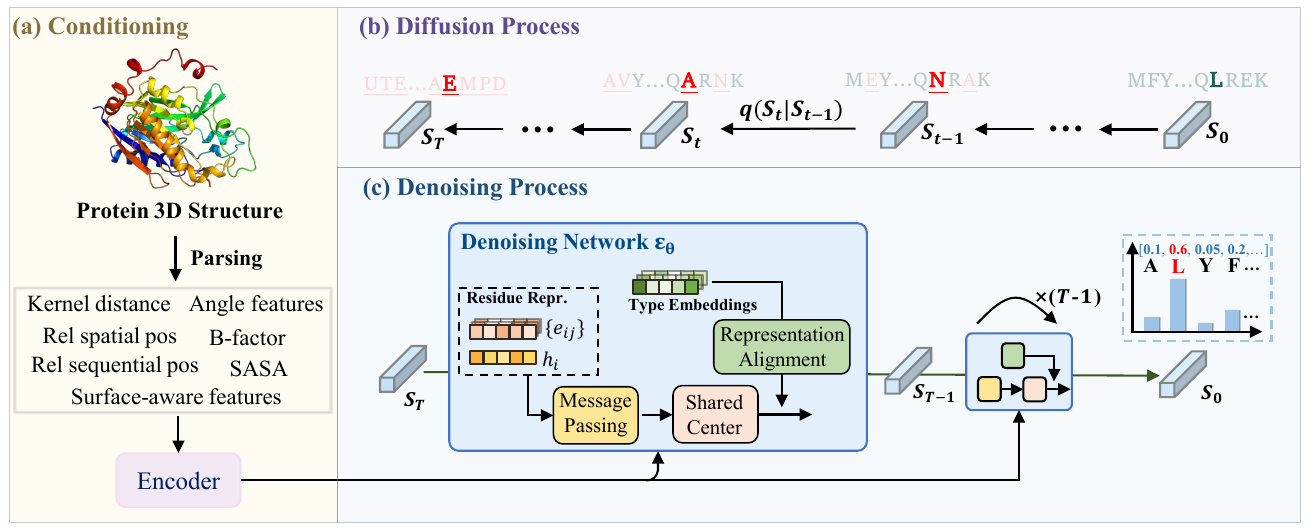}
    \vspace{-6mm}
    \caption{\small Overview illustration of our $\Ours$ model.
    We take a certain amino acid type in the sequence as an example (highlighted in the figure) to illustrate the process of the model. In the diffusion process, the amino acid type is represented by characters, where green characters represent the correct amino acid type and underlined red characters represent the incorrect amino acid type after transfer.}
    \label{fig1}
    \vspace{-4mm}
\end{figure*}
\section{Method}
In this section, we introduce our novel method, $\Ours$, to use the diffusion model with representation alignment for protein inverse folding. As shown in Figure~\ref{fig1}, Our approach starts with protein structure compression and diffusion modeling. We then delve into our denoising network, comprising message passing with cell module, shared center, and representation alignment method.

\subsection{Protein Structure Compression}
To compress protein 3D structure into a compact representation, we parse the backbone structure and construct a K-nearest neighbor graph $\gG(\mX, \mE)$ based on the coordinates of $\mC_\alpha$ atoms, where K is 30 at default.
The $\gG(\mX, \mE)$ comprises node features $\mX \in \R^{N, d_n}$ and edge features $\mE \in \R^{M, d_e}$, where these features are used to represent residues and their relationships, and $N$ and $M$ denote the numbers of nodes and edges, respectively.
Following~\citet{YiZ0L023}, node and edge features are defined as follows:
\begin{align}
    \mX &= \encodernode(\mX_b; \mX_{sasa}; \mX_a; \mX_s)\\
    \mE &= \encoderedge(\mE_k; \mE_{sp}; \mE_{se})
\end{align}
where B-Factor $\mX_b \in \R^{N,1}$ and solvent-accessible surface area (SASA) $\mX_{sasa} \in \R^{N,1}$ are derived from the scalar values of $\mC_\alpha$ atoms. B-Factor reflects the static stability of the protein, while SASA provides insights into protein folding and hydrophobicity. 
Angle features $\mX_a \in \R^{N,4}$ contain the sine and cosine of backbone dihedral angles $\psi$ and $\phi$, i.e., local geometry of residues. Surface-aware features $\mX_s \in \R^{N,5}$ are encoded as vectors according to a set of hyperparameters $\lambda$, representing the normalized distances between the central amino acid and its one-hop neighbors~\cite{GaneaHBBBJ022}. 
For edge features, kernel-based distances $\mE_k \in \R^{M, 15}$ are described using Gaussian radial basis functions (RBF) with varying bandwidths to capture distance information between connected residues at different scales, totaling 15 different distance features. 
$\mE_{sp} \in \R^{M, 12}$ are derived from the heavy atom positions of the corresponding residues, totaling 12 relative position features~\cite{YiZ0L023}. 
The relative sequence distances $\mE_{se} \in \R^{M, 66}$ use 65-dimensional one-hot vectors as bins to encode the relative sequence distance of two residues in the protein chain, along with a binary feature indicating whether the Euclidean distance between two connected residues is less than a specified threshold.

\subsection{Diffusion Modeling}
Our method is based on a diffusion modeling framework~\cite{austin2021structured} for protein inverse folding, which includes both diffusion and denoising processes.

\paragraph{Diffusion Process}
In the diffusion process, noise is introduced to the node's AA type. Specifically, at timestep $t$, each node's AA type $\mS_0$ in the sequence transforms to other amino acid types using a probability transfer matrix~\cite{hoogeboom2021argmax, austin2021structured} $\mQ_t=\alpha_t\mI+(1-\alpha_t)\1_d\1_d^\top/d, \mQ_t \in \R^{20 \times 20}$ with $\mI$ being the transpose of the identity matrix and $d$ being the number of AA types and $\1$ being the one vector of dimension $d$, i.e.,
\begin{align}
    p\left(\mS_t\middle| \mS_{t-1}\right) = \mS_{t-1} \cdot \mQ_t
\end{align}
where $\mS_t$ and $\mS_{t-1}$ represent node's noise AA type in step $t$ and $t-1$, respectively.
Similar to DDPM~\cite{ho2020denoising}, we can compute node's noise AA type in step $t$ from initial step, denoted as follows:
\begin{align}
    p\left(\mS_t\middle| \mS_0\right) = \mS_0\cdot \bar{\mQ}_t\ 
\end{align}
$\bar{\mQ}_t$ denotes transition probability from initial step to $t$ directly, and $\mS_0$ represents node's original AA type.

\paragraph{Denoising Process}
In the denoising process, each node's noise AA type is sampled from the uniformly prior distribution and iterated back to the initial distribution.
The transformation between distributions is sketched as follows:
\begin{align}
    p_\theta\left(\mS_{t-1}\middle| \mS_t,\gG\right)\!=\!\sum_{\hat{s}_0}{q(\mS_{t-1}|{\hat \mS}_0,\mS_t,\gG)p_\theta({\hat \mS}_0|\mS_t,\gG)}\!
\end{align}
where ${\hat \mS}_0$ is 
predicted AA type and $q(\mS_{t-1}|{\hat \mS}_0,\mS_t, \gG)$ represent posterior that can be computed as follows:
\begin{align}
q(\mS_{t-1}|{\hat \mS}_0,\mS_t, \gG)\!=\!\mathbf{DIST}(\mS_{t-1}|\frac{\normalsize \mS_t\scriptsize{\mQ_t^{T}}\!\odot\!\normalsize {\hat \mS}_0 \scriptsize{\bar \mQ_{t-1}^{T}}}{{\hat \mS}_0\scriptsize{\bar \mQ_t}\normalsize{\mS_t^T}})
\end{align}
where $\mathbf{DIST}$ is a categorical distribution over 20 AA types with probabilities computed by the posterior distribution.

\subsection{$\Ours$ Denoising Network}
As shown in Figure~\ref{fig1}, we propose a $\Ours$ denoising network $\eps_\theta(\mS_t, t, \gG)$ to predict $p_\theta({\hat \mS}_0|\mS_t,\gG)$.
The network includes three key components: the Message Passing module, the Shared Center module, and the Representation Alignment module. 
Initially, we concatenate the corresponding $\mS_t$ and $\mX$ to form the initial node representation, $\vh=\{\vh_1,...\vh_i,...\vh_N\}$.

\paragraph{Message Passing.}
The Message Passing module updates node representations using information from neighboring nodes and their relationships. Firstly, given a node $\vh_i$ as an example, a gating mechanism within the $\mathbf{Cell}$ in Eq.(\ref{c1}-\ref{c11}), dynamically adjusts both node and edge features, producing the message $\vm'_{ij}$. Specifically, the node $\vh_i$ is concatenated with its neighboring node $\vh_j$ as the message $\vm_{ij}$, which is then merged with the edge features $\ve_{ij}$ as gates, i.e.,
\begin{align}
\label{c1}
\vg^{(1)}_{ij} & = \sigma(\Linear([\ve_{ij}; \vm_{ij}])\\
\vg^{(2)}_{ij} & = \sigma(\Linear([\ve_{ij}; \vm_{ij}]))
\end{align}
where $\sigma$ is the sigmoid function. $\vg^{(1)}_{ij}$ and $\vg^{(2)}_{ij}$ are two gates, which are used to update message $\vm_{ij}$, i.e,
\begin{align}
\vn_{ij} & = \mathbf{Act}(\Linear(\ve_{ij})+\vg^{(1)}_{ij}\odot \Linear(\vm_{ij})) \\
\vm'_{ij} \!&= \vg^{(2)}_{ij}\odot \vm_{ij}+(1-\vg^{(2)}_{ij})\odot \vn_{ij}
\label{c11}
\end{align}
where $\mathbf{Act}$ is activation function.
Subsequently, messages from all neighboring nodes are aggregated to update the central node's representation, i.e.,
\begin{align}
\vh_i^\prime=\mathbf{MLP}(\vh_i,\sum_{j\in\gN_i} \vm'_{ij}),
\end{align}
where $\vh_i^\prime$ is the updated feature of node $i$, and $\gN_i$ represents the set of neighbors of node $i$.
\begin{figure}[!t]
    \centering
    \includegraphics[width=\linewidth]{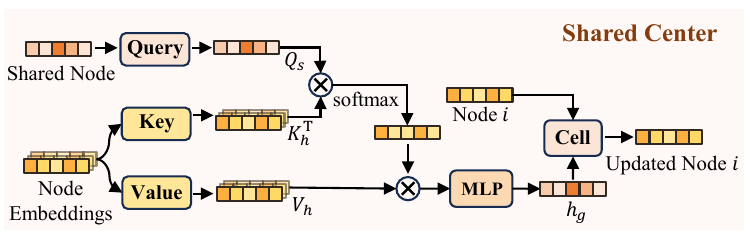}
    \vspace{-6mm}
    \caption{Details of the Shared Center module in the $\Ours$ model, where the shared node is initialized to aggregate contextual information of the entire protein chain, the $\mathbf{Cell}$ module is specifically formulated as Eq.(\ref{c2}-\ref{c3}).}
    \label{SC}
    \vspace{-4mm}
\end{figure}
\paragraph{Shared Center.}
To effectively enhance representations for residues, we propose the Shared Center module, as shown in Figure~\ref{SC}, which integrates the contextual information of the entire protein chain as shared content and selectively distributes it to each residue for access. Specifically, a virtual shared node $\vh_{s}\in \R^{1, d}$ is initialized at first. Subsequently, it is transformed to the query space, while the node representations aggregated from the Message Passing are transformed to the key and value spaces, i.e,
\begin{align}
    \mQ_s=\mW_q^s \cdot \vh_s;~~\mK_h = \mW_k^h \cdot \vh^\prime; ~~\mV_h = \mW_v^h \cdot \vh^\prime
\end{align}
where $\mW^s_q\in \R^{d, d}$, $\mW_k^h\in \R^{d, d}$, $\mW_v^h\in \R^{d, d}$ are the projection matrices.
$\vh_{s}$ is randomly initialized, and $\vh^\prime = \{\vh_1^\prime, \vh_2^\prime, \dots, \vh_n^\prime\}$ represents all nodes in the protein. 
We compute the attention score between them and use the score to adaptively aggregate information from the entire protein chain, i.e,
\begin{align}
    \vh_g &= \mathbf{Softmax}\left(\frac{\mQ_s \cdot \mK_h^\mathrm{T}}{\sqrt{\vd}}\right)\mV_h
\end{align}
where the output, $\vh_g$, encapsulates the contextual information of the entire protein structure, which is then provided to each residue for access. 
We employ the $\mathbf{Cell}$ module with a gating mechanism shown in Eq.(\ref{c2}-\ref{c3}), which selectively receives the quantity of information based on the current node to enhance residue representation, i.e., 
\begin{align}
\label{c2}
\vg^{(1)}_i & = \sigma(\Linear([\vh'_i;~ \vh_g])\\
\vg^{(2)}_i & = \sigma(\Linear([\vh'_i;~ \vh_g]))
\end{align}
where $\sigma$ is the sigmoid function. $\vg^{(1)}_i$ and $\vg^{(2)}_i$ are two gates, which are used to receive information from the central contextual feature $\vh_g$ according to node $\vh_i^\prime$, i.e,
\begin{align}
\vc_i & = \mathbf{Act}(\Linear(\vh_i^\prime)+\vg^{(1)}_i\odot \Linear(\vh_g)) \\
\label{c3}
\widetilde{\vh}_i \!&= \vg^{(2)}_i\odot \vh_g+(1-\vg^{(2)}_i)\odot \vc_i
\end{align}
where $\widetilde{\vh}_i$ is the updated feature of node $i$.
\begin{figure}[!t]
    \centering
    \includegraphics[width=\linewidth]{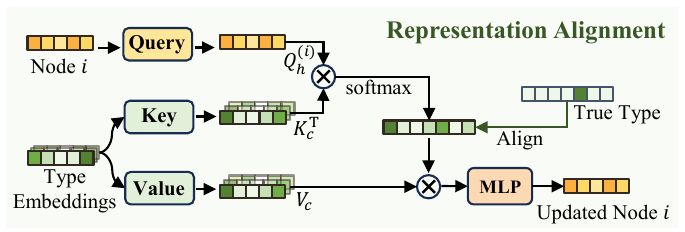}
    \vspace{-6mm}
    \caption{Details of the Representation Alignment module in the $\Ours$ model. The true type is used to enforce the residue to align with the clean type embeddings during the denoising process.}
    \label{RA}
    \vspace{-3mm}
\end{figure}
\paragraph{Representation Alignment.}
To normalize residue representations during the denoising process, we introduce the representation alignment module, in Figure~\ref{RA}. This module adaptively injects semantic embeddings of AA type into the noisy hidden representations. 
Firstly, all types are initialized as $\vh_c \in \R^{20, d}$. They are then mapped to the key and value spaces served as a reference, while the node representation $\widetilde{\vh}_i$ is mapped to the query space served as a request, i.e.,
\begin{align}
    \mQ_h^{(i)}=\mW_q \cdot \widetilde{\vh}_i;~~\mK_c = \mW_k \cdot \vh_c; ~~\mV_c = \mW_v \cdot \vh_c
\end{align}
where $\mW_q \in \R^{d, d}$, $\mW_k \in \R^{d, d}$, and $\mW_v \in \R^{d, d}$ are the projection matrices.
We calculate the correlation between the node and the 20 AA types using scaled dot-product attention, i.e.,
\begin{align}
\vp(h^{(i)}) &= \mathbf{Softmax}\left(\frac{\mQ_h^{(i)} \cdot \mK_c^\mathrm{T}}{\sqrt{\vd}}\right)
\end{align}
where $\vp(\vh^{(i)}) \in \R^{1,20}$ represents the correlation between the $i$-th node and the 20 AA types. The AA-type semantic embeddings are then weighted and normalize the node, i.e.,
\begin{align}
\vh_i^l = \vp(\vh^{(i)}) \cdot \mV_c
\end{align}
To ensure each node aligns with the corresponding AA type, we apply cross-entropy loss to constrain the correlation matrix, i.e.,
\begin{align}
\gL_{attn} = - \frac{1}{N}\sum_{i=1}^{N} {\vp(h^{(i)}_{true}) \log(\vp(h^{(i)}))}
\end{align}
where $N$ is the number of nodes, and $\vp(h^{(i)}_{true})$ is ground truth AA types of node $i$. Finally, obtained from the share center and representation alignment modules, the node representations are enhanced in each layer.

Following the diffusion framework~\cite{peebles2023scalable}, The time step $t$ is mapped to $\gamma$ and $\beta$ to dynamically adjust the scale of features after computing the representation at each layer:
\begin{align}
\vh_i^l &= \vh_i^l \ast (\gamma + 1) + \beta
\end{align}
$\gamma$ and $\beta$ denote scale and shift respectively. The dimensions of them are consistent with the node representation $\vh_i^l$. The final node representation in layer $m$ is mapped to the 20 AA types, which are associated with the secondary structure embedding $ss$~\cite{YiZ0L023}:
\begin{align}
\vp_i &= \mathbf{MLP}(\vh_i^m + \mathbf{Linear}(ss))
\end{align}
where $\vp_i \in \R^{20}$ represents the predicted amino acid type of the $i$-th node.

\subsection{Training Objective}
For the model training, we employed cross-entropy loss to optimize the model's final predictions for each node type, i.e.,
\begin{align}
\gL_{pred} &= - \frac{1}{N}\sum_{i=1}^{N} \vp(h^{(i)}_{true}) \log(\vp_{i}) \\
\gL  &= \alpha \cdot \gL_{pred} + \lambda  \cdot \gL_{attn}
\end{align}
where $\alpha$ and $\lambda$ are weight coefficients, and $\vp(h^{(i)}_{true})$ represents the true type of the $i$-th residue, $\vp_i$ represents the predicted type of the $i$-th residue and $\gL$ represents the final loss, which includes the prediction cross-entropy loss $\gL_{pred}$ and the constraint loss $\gL_{attn}$ in representation alignment module.

\section{Experiments}

\subsection{Dataset and Evaluation Metrics}
In our experiments, we compare our method against other approaches on the CATH4.2 dataset, a widely-used benchmark categorized based on the CATH topology classification~\cite{orengo1997cath}. Following the data-splitting in previous works, e.g., GraphTrans~\cite{ingraham2019generative}, PiFold~\cite{Gao0L23}, and GRADE-IF~\cite{YiZ0L023}, we divide the dataset into 18,024 proteins for training, 608 proteins for validation, and 1,120 proteins for testing. In addition, we extend our evaluation to the TS50 and TS500 datasets to validate the generalization capability of our model. 
We employ two evaluation metrics for assessing generated AA sequences, i.e., \textbf{Recovery} rate and \textbf{Perplexity}. 
The recovery rate quantifies the accuracy of the generated sequences compared to the ground truth, providing insight into the model's precision. 
Perplexity measures the uncertainty in the model's predictions, reflecting its confidence and ability to generalize to unseen data.

\subsection{Experimental Setting}
To comprehensively evaluate the model's performance to recover sequences, we divide the test data into three categories: ``short'', ``single'', and ``all'', as shown in Table \ref{main}. 
The ``short'' comprises proteins with AA sequence lengths fewer than 100, and ``single'' includes proteins composed of a single chain; ``all'' encompasses the entire test dataset. 
The denoising network consists of six stacked layers, and the timestep for the diffusion model is set to 500. 
The model is trained for a total of 70,000 steps with a batch size of 32 and gradient accumulation over two steps on an NVIDIA A6000 GPU. 
We employ the Adam optimizer with a learning rate of 0.0005 and a weight decay of 0.00001.
The weight of $\alpha$ and $\lambda$ are both 0.5. 
In the inference process, we utilize accelerated inference methods based on~\cite{YiZ0L023,song2020denoising}, with a skip interval of 500 and a single denoising step, striking a balance between recovery rate and perplexity.

\subsection{Main Results}
\begin{table}[!t]\small
\caption{Experiment result on the CATH4.2 dataset.}
\label{main}
\centering
\setlength{\tabcolsep}{3pt}
\begin{tabular}{lcclccc}
\toprule
\multicolumn{1}{l}{\multirow{2}{*}{\textbf{Method}}} & \multicolumn{3}{c}{\textbf{Perplexity↓}} & \multicolumn{3}{c}{\textbf{Recovery(\%)↑}} \\ \cmidrule(r){2-4}\cmidrule(r){5-7} 
\multicolumn{1}{c}{}& \textbf{Short}& \textbf{Single}& \textbf{All}& \textbf{Short}& \textbf{Single}& \textbf{All}\\ \midrule
StructGNN~\cite{ingraham2019generative}    & 8.29& 8.74& 6.40& 29.44& 28.26& 35.91    \\
GraphTrans~\cite{ingraham2019generative}   & 8.39& 8.83& 6.63& 28.14& 28.46& 35.82    \\
GCA~\cite{GCA22}          & 7.09& 7.49& 6.05& 32.62& 31.10& 37.64    \\
GVP~\cite{GVP21JingESTD} & 7.23& 7.84& 5.36& 30.60& 28.95& 39.47    \\
AlphaDesign~\cite{2022alphadesign}  & 7.32& 7.63& 6.30& 34.16& 32.66& 41.31    \\
ProteinMPNN~\cite{dauparas2022robust}  & 6.21& 6.68& 4.61& 36.35& 34.43& 45.96    \\
PiFold~\cite{Gao0L23}  & 6.04& 6.31& 4.55& 39.84& 38.53& 51.66    \\
GRADE-IF~\cite{YiZ0L023}     & 5.49& 6.21& 4.35& 45.27& 42.77& 52.21    \\
VFN-IF~\cite{VFN}       & 5.70& 5.86& 4.17& 41.34& 40.98& 54.74    \\
\rowcolor{gray!15}$\Ours$ (ours)& \textbf{4.06}& \textbf{4.76}& \textbf{2.93}& \textbf{53.57}& \textbf{48.95}& \textbf{64.07} \\\hline\midrule
\multicolumn{7}{c}{\textit{w/ External Knowledge}}  \\\midrule
LM-Design~\cite{ZhengDX0YG23}        & 6.77 & 6.46 & 4.52 & 37.88 & 42.47 & 55.65 \\
KW-Design~\cite{gao2023KD} & 5.48 & 5.16 & 3.46 & 44.66 & 45.45 & 60.77 \\
\bottomrule
\end{tabular}
\end{table}
To validate the effectiveness of our method, we compared it with other strong competitors using the CATH4.2 benchmark, and the results are shown in Table~\ref{main}.
Experimental results demonstrate that our model achieves state-of-the-art performance in AA sequence recovery and perplexity. To the best of our knowledge, our method is the first to achieve 60\% recovery without external knowledge of pre-trained language models. In addition, compared to the VFN-IF model, our approach improves the recovery rate by 9.33\%, confirming its superior performance. Compared with GRADE-IF, our method increases the recovery rate by 11.86\%, indicating the effectiveness of representation alignment in the denoising network for sequence recovery. Furthermore, while the LM-Design and KW-Design models utilize external knowledge from the pre-trained ESM~\cite{esm2023}, achieving recovery rates of 55.65\% and 60.77\%, respectively, our model improves the recovery rates by 8.42\% and 3.30\%. This demonstrates that our model delivers strong sequence recovery capabilities without external knowledge, thereby reducing computational complexity during the inference stage.

\subsection{Generalization Capability Analysis}
\begin{table}[t]\small
\centering
\caption{Results of experiments on the TS50 and TS500 datasets. PPL refers to Perplexity, and Rec indicates Recovery (\%).}
\label{ts50}
\setlength{\tabcolsep}{9pt}
\begin{tabular}{lcccc}
\toprule
& \multicolumn{2}{c}{\textbf{TS50}} & \multicolumn{2}{c}{\textbf{TS500}} \\ 
\cmidrule(r){2-3}\cmidrule(r){4-5}
\multirow{-2}{*}{\textbf{Method}} & \textbf{PPL}& \textbf{Rec}& \textbf{PPL} & \textbf{Rec}\\ 
\midrule
StructGNN~\cite{ingraham2019generative}  & 5.40 & 43.89 & 4.98 & 45.69\\
GraphTrans~\cite{ingraham2019generative} & 5.60 & 42.20 & 5.16 & 44.66\\
GVP~\cite{GVP21JingESTD}        & 4.71 & 44.14 & 4.20 & 49.14\\
GCA~\cite{GCA22}        & 5.09 & 47.02 & 4.72 & 47.74\\
AlphaDesign~\cite{2022alphadesign}& 5.25 & 48.36 & 4.93 & 49.23\\
ProteinMPNN~\cite{dauparas2022robust}& 3.93 & 54.43 & 3.53 & 58.08\\
PiFold~\cite{Gao0L23}     & 3.86 & 58.72 & 3.44 & 60.42\\
GRADE-IF~\cite{YiZ0L023}   & 3.71 & 56.32 & 3.23 & 61.22\\
VFN-IF~\cite{VFN}     & 3.58 & 59.54 & 3.19 & 63.65\\
\rowcolor{gray!15}$\Ours$ (ours) & \textbf{2.67} & \textbf{67.03} & \textbf{2.31}& \textbf{71.61}\\\hline\midrule
\multicolumn{5}{c}{\textit{w/ External Knowledge}}  \\\midrule
LM-Design~\cite{ZhengDX0YG23}  & 3.50 & 57.89 & 3.19 & 67.78 \\
KW-Design~\cite{gao2023KD}  & 3.10 & 62.79 & 2.86 & 69.19 \\
\bottomrule
\end{tabular}
\end{table}
To verify the generalization capability of our model, we directly evaluated the trained model on the TS50 and TS500 datasets. 
The TS50 and TS500 datasets consist of 50 and 500 test proteins, respectively. 
As shown in Table~\ref{ts50}, our model achieves state-of-the-art performance on both datasets, significantly outperforming existing methods. 
Specifically, our model achieves a perplexity (PPL) of 2.67 on TS50 and 2.31 on TS500, substantially outperforming existing approaches such as GRADE-IF and VFN-IF. For recovery rate (Rec), our model achieves 67.03\% on TS50 and 71.61\% on TS500. Notably, it is the first model, to our knowledge, that exceeds a recovery rate of 70\% on the TS500 and over 65\% on the TS50 without leveraging external knowledge in training.
These results underscore the robustness and generalization capability of our approach. While models incorporating external knowledge in training, such as LM-Design~\cite{ZhengDX0YG23} and KW-Design~\cite{gao2023KD}, also achieve competitive results, our model demonstrates that better performance can be reached purely through representation alignment in denoising network, thus reducing the reliance on external domain-specific information.

\subsection{Ablation Study}
\begin{table}[t]\small
\caption{Ablation study. ``w/o RA'' indicates the model without representation alignment module, ``w/o SC'' refers to the model without shared center module, and ``w/o ALL'' denotes the model without SC and RA and the cell module in message passing.}
\vspace{-1mm}
\centering
\setlength{\tabcolsep}{5pt}
\label{ablation}
\begin{tabular}{lcccccc}
\toprule
\multicolumn{1}{l}{\multirow{2}{*}{\textbf{Model}}} & \multicolumn{2}{c}{\textbf{CATH}} & \multicolumn{2}{c}{\textbf{TS50}} & \multicolumn{2}{c}{\textbf{TS500}} \\ \cmidrule(r){2-3}\cmidrule(r){4-5}\cmidrule(r){6-7}
\multicolumn{1}{c}{} & \textbf{Rec}   & \textbf{PPL}     & \textbf{Rec}   & \textbf{PPL}     & \textbf{Rec}   & \textbf{PPL}   \\ \midrule
$\Ours$                               & \textbf{64.07}  & \textbf{2.93}  & \textbf{67.03}  & \textbf{2.67}  & \textbf{71.61}  & \textbf{2.31}  \\\midrule
w/o RA           & 63.13 & 3.00 & 64.46 & 2.80 & 70.32 & 2.39 \\
w/o SC               & 61.60 & 3.16 & 64.24 & 2.89 & 68.74 & 2.52 \\
w/o SC \& RA & 61.48 & 3.17 & 63.68 & 2.92 & 68.73 & 2.51 \\ 
w/o ALL &60.96 &3.21 & 63.61 & 2.95 & 68.36 & 2.54 \\ 
\bottomrule
\end{tabular}
\vspace{-3mm}
\end{table}
To evaluate the impact of each module within the denoising network, we conduct an ablation study, and the results are shown in Table~\ref{ablation}. The performance metrics are evaluated across three datasets: CATH, TS50, and TS500. 
Firstly, removing the representation alignment module (``w/o RA'') disrupts the model's understanding of various residue types, leading to a decline in performance across the CATH, TS50, and TS500 datasets. It demonstrates the necessity of the model's understanding of various residue types by aligning their representation during the denoising process. 
Similarly, excluding the shared center module (``w/o SC'') also leads to a marked decline in performance. Without this module, the model is restricted to relying purely on graph neural network (GNN)-based neighbor inter-residue interactions, without leveraging holistic information from the entire protein chain. This limitation hinders the model's ability to contextualize residue interactions, as evidenced by decreased recall and increased perplexity across all datasets. These results confirm the effectiveness of integrating global chain-level information to enrich residue representations and improve predictive accuracy.
When both the representation alignment and shared center modules are simultaneously removed (``w/o RA \& SC''), the model suffers further performance degradation. This reinforces the complementary contributions of these two components, highlighting that both are indispensable for capturing complex residue dependencies within the denoising network.
Lastly, we explore the role of the Cell module within the Message Passing part, which integrates node and edge representations from neighboring nodes. When the Cell module is replaced with a MLP, model performance declines, indicating the crucial role of the Cell module in effectively integrating neighboring node and edge representations.

\subsection{Nonlinear Analysis in Shared Center}
\begin{figure}[t]
\centering
\includegraphics[width=\linewidth]{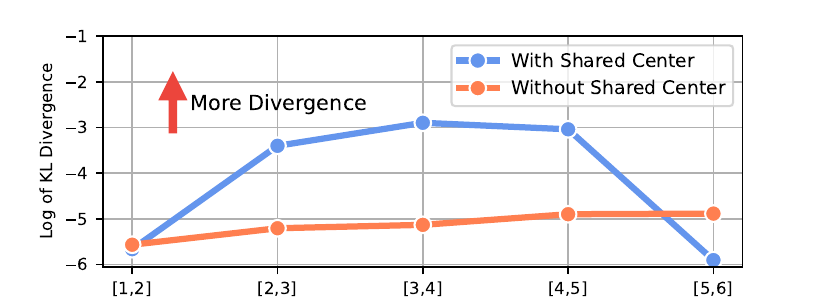}
\caption{\small Nonlinear features analysis of shared center module on layer output. The x-axis shows two adjacent layers, and the y-axis represents the logarithm of KL divergence, which measures changes in node feature distributions. The polyline depicts nonlinear divergences in node representations for 50 randomly selected protein cases from CATH4.2 test set with or without the shared center module. Arrows indicate the direction towards nodes with stronger nonlinear features.}
\label{plot_global}
\vspace{-4mm}
\end{figure}
To better understand the effectiveness of the Shared Center (SC) module, we examine its influence on the nonlinear characteristics of layer outputs. In Figure~\ref{plot_global}, we compare two scenarios: one where the SC module is used and another where it is not. The y-axis represents the logarithm of the KL divergence, which quantifies the changes in feature distributions between adjacent layers. A higher value indicates a greater divergence, suggesting more pronounced nonlinear transformations.
The results reveal that incorporating the SC module significantly increases the KL divergence across layers, especially in the earlier stages. This indicates that SC enhances the model's ability to capture and propagate complex nonlinear patterns, thereby improving its representation of inter-residue relationships. In contrast, without the SC module, the divergence remains consistently lower, suggesting limited capacity for nonlinear feature extraction. Thus, the SC module's impact is particularly beneficial for tasks that require nuanced representation of protein structures.

\subsection{Visualization of Representation Alignment}
\begin{figure}[!t]
\centering
\includegraphics[width=\linewidth]{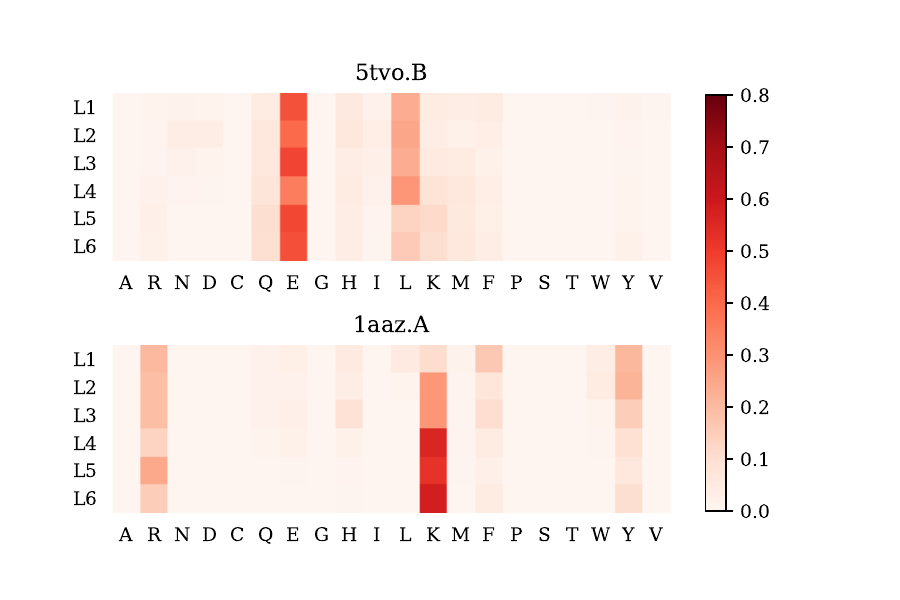}
\caption{\small Attention visualization of a single residue's alignment across different layers in the denoising network.
The horizontal axis represents 20 types of amino acids, and {L1-L6} on the vertical axis represents the $i^{th}$ layer in the denoising network, respectively.
5tvo.B and 1aaz.A represent two proteins selected from the CATH test set randomly.
}
\label{fig3}
\vspace{-3mm}
\end{figure}
\begin{figure}[!t]
    \centering    
    \includegraphics[width=\linewidth]{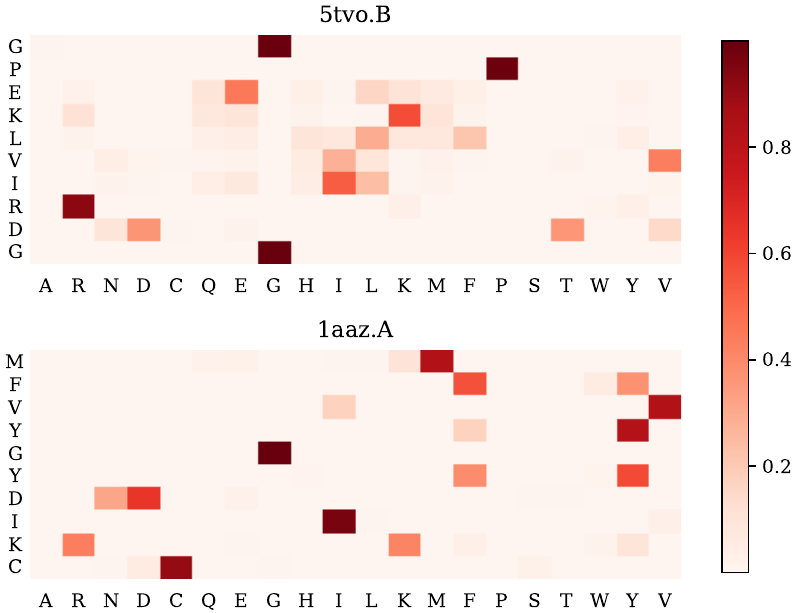}
    \caption{\small Attention visualization of multiple residues in the final layer. The horizontal axis represents 20 types of amino acids, and each character in the vertical axis represents the predicted AA type of randomly selected residue.}
    \label{fig4}
\vspace{-3mm}
\end{figure}
To investigate representation alignment, we present visualization results of model predictions. Specifically, we analyze the AA types to which individual residues attend across different layers of the denoising network, as shown in Figure~\ref{fig3}. We also show the AA types attended to by multiple residues in the final layer, in Figure~\ref{fig4}.
In Figure~\ref{fig3}, the vertical axis (L1 to L6) represents the network layers, while the horizontal axis represents the 20 AA types. 
The values indicate the attention weights computed by the representation alignment module for the specific residue and each of the 20 AA types.
For the 5tvo.B protein, the true type of the randomly chosen residue is glutamic acid (E), and for the 1aaz.A protein, it is lysine (K).
The results demonstrate that as the number of layers increases, the attention weight for the node corresponding to the true AA type of each residue gradually rises. By the final layer, the residue aligns with its true AA type, suggesting that the model effectively aligns residues with their true AA types and incorporates this information into the node representations.
In addition, Figure~\ref{fig4} visualizes the attention alignment of multiple residues in the model's final layer. These residues are randomly selected from those with correctly predicted types.
The horizontal axis represents the 20 AA types, while the vertical axis represents the true amino acid types of the correctly predicted residues. The visualization shows that the correctly predicted residues have the highest attention weights for their true AA types, which indicates that injecting true AA-type information into the residue representations is beneficial for prediction.

\subsection{Folding Ability}
\begin{figure}
\centering
\includegraphics[width=1.0\linewidth]{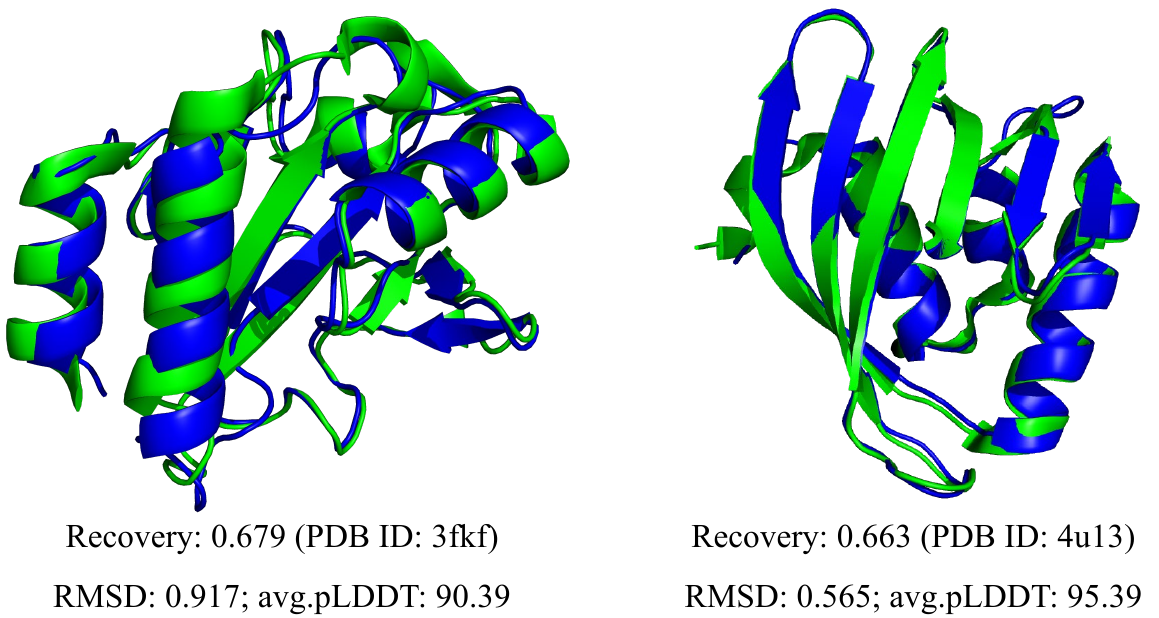}
\caption{\small Comparison of the folding between predicted (Blue) and native (Green) structures, where the predicted structures are generated using AlphaFold2 based on $\Ours$-designed AA sequences.}
\label{3fkf}
\vspace{-3mm}
\end{figure}
We further explore the folding ability of the generated amino acid sequences to verify its rationality. Specifically, we randomly select test proteins 3fkf and 4u13 from the CATH4.2 test set and utilize the protein structure prediction method ColabFold~\cite{mirdita2022colabfold}, which offers user-friendly access to AlphaFold2~\cite{jumper2021highly} for predicting the 3D structures of the generated amino acid sequences. These predicted structures are then aligned with the corresponding PDB structures. As shown in Figure~\ref{3fkf}, the recovery rate of the generated sequences 3fkf is 0.679, and secondary structure elements such as $\alpha$-helices and $\beta$-sheets are effectively formed. The average pLDDT score is 90.39, and the RMSD is 0.917, where the average pLDDT score assesses confidence in the predicted structure, and the RMSD measures the deviation between the predicted and fixed structures. These results demonstrate the validity and rationality of our model in generating new sequences based on fixed backbone structures.

\section{Conclusion}
In this paper, we propose a novel approach $\Ours$ that leverages diffusion models with representation alignment for protein inverse folding. Firstly, our method integrates contextual information from the entire 3D structure as shared center and assigns it selectively to each residue to maintain inter-residue communication, enhancing the protein representation learning. Moreover, we normalize residue representations and align noisy hidden representations with clean type-specific feature embeddings in the denoising process. In addition, our cell module effectively decouples and computes the relevance of adjacent node and edge information. In experiments, our method surpasses existing leading approaches on the CATH4.2, TS50 and TS500 datasets.

{
    \small
    \bibliographystyle{ieeenat_fullname}
    \bibliography{main}
}

\clearpage
\appendix
% \setcounter{page}{1}
% \maketitlesupplementary

\section{Visualization Cases of Representation Alignment}
\begin{figure*}[t]
    \centering
    \includegraphics[width=0.93\linewidth]{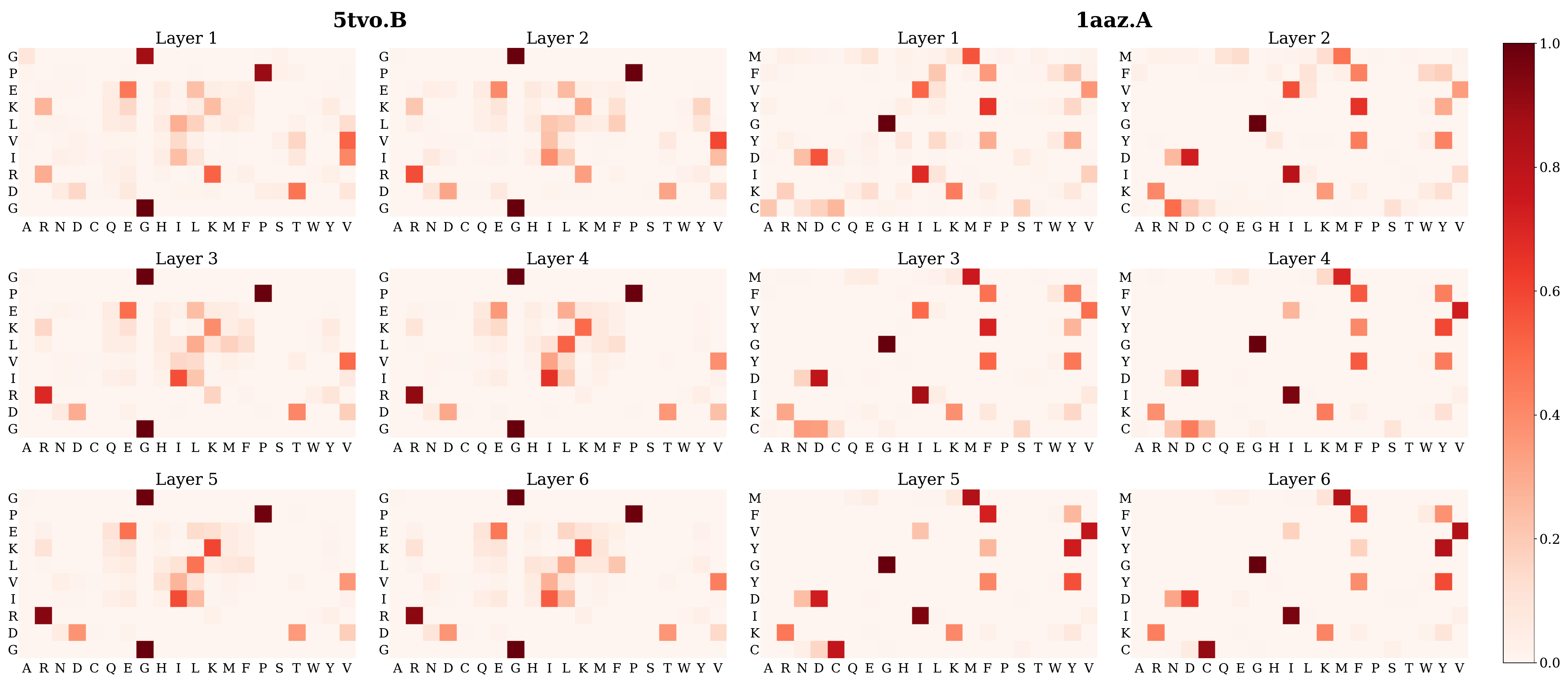}
    \includegraphics[width=0.93\linewidth]{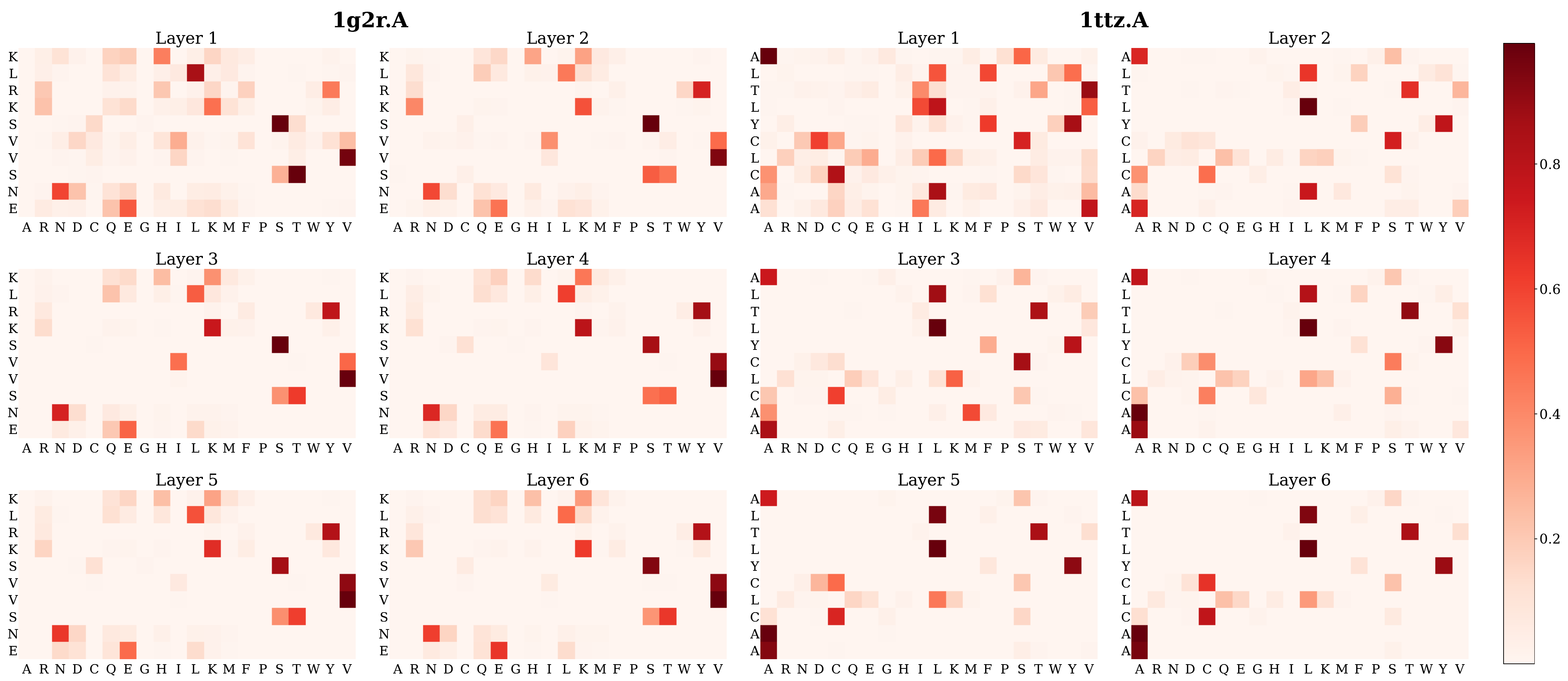}
    \includegraphics[width=0.93\linewidth]{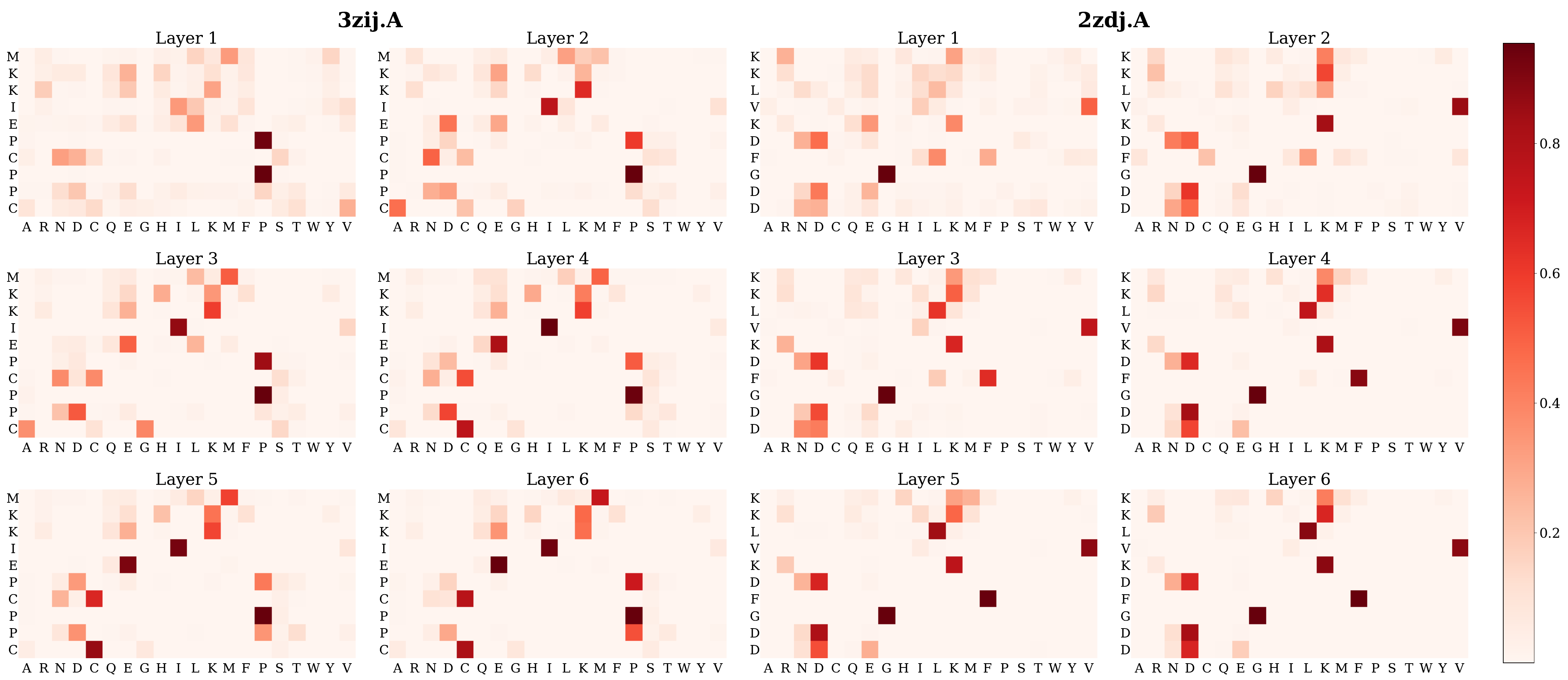}
    \caption{ \small Attention visualization across the representation alignment module in different layers of the model, where multiple residues are selected randomly as cases to show their alignment with the clean type embeddings.}
    \label{fig5}
\end{figure*}
To deeply verify the effectiveness of the representation alignment module, we perform detailed visualization and analysis on more cases randomly selected from the CATH test set. 
As shown in Figure~\ref{fig5}, We randomly select six protein cases and randomly choose ten residue positions from each case's sequence to visualize the attention weights of the representation alignment module across different layers (a total of six layers) in the $\Ours$ denoising network. The horizontal axis represents the twenty amino acid types, while the vertical axis indicates the amino acid types predicted by the model at these ten positions. The results demonstrate that as the network depth increases, the semantic feedback of type embeddings to residue representations becomes progressively more pronounced. The representation alignment module significantly normalizes the ability of residue representations to correctly distinguish between the semantic features of different amino acid types, thereby ensuring more precise representation alignment for amino acid types.

\section{Visualization Cases of Folding Ability}
\begin{figure*}[t]
    \centering
    \includegraphics[width=0.93\linewidth]{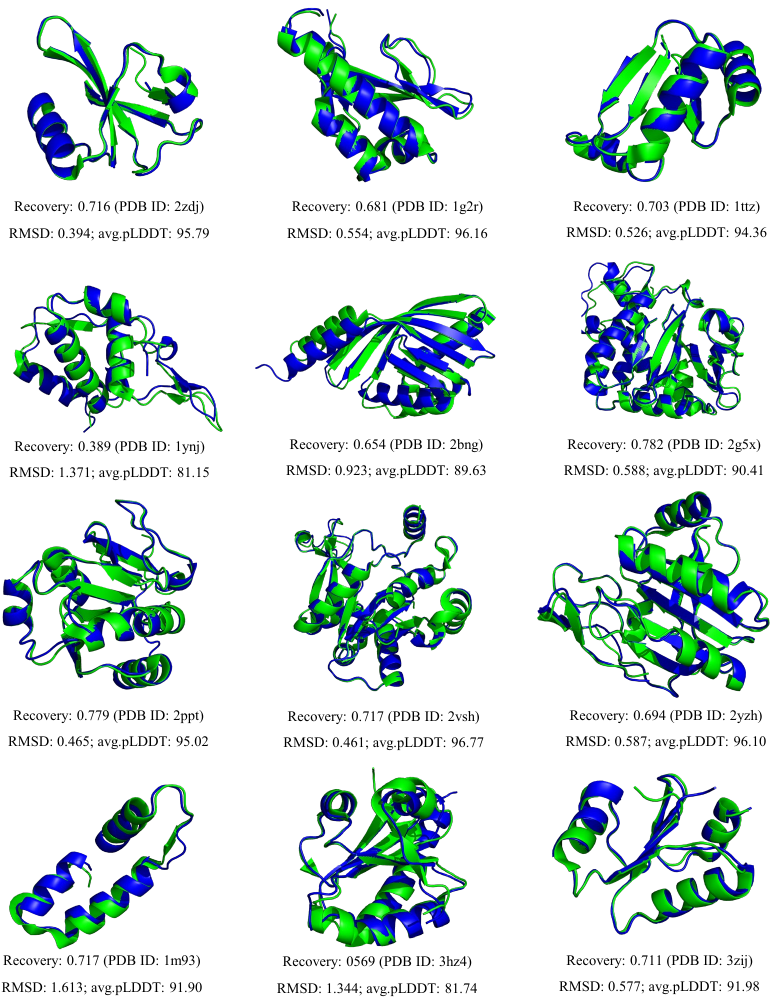}
    \caption{\small More visualization cases of comparison of predicted(Blue) structures folded by $\Ours$-designed AA sequences and native(Green) structures, where the cases are selected from the CATH test dataset randomly.}
    \label{fig6}
\end{figure*}
We select additional protein cases from the CATH test set and design amino acid sequences based on their backbone structures using our proposed method. We then employ the widely recognized structural prediction tool ColabFold, which offers user-friendly access to AlphaFold2, to fold  $\Ours$-designed sequences into their corresponding tertiary structures. This approach allows us to evaluate both the folding ability of the designed sequences and the structural plausibility of the resulting protein tertiary structures. As illustrated in Figure~\ref{fig6}, the rationality of these cases serves as validation for the effectiveness of our method in protein reverse folding.

\end{document}